\documentclass{article}

\usepackage{arxiv}

\usepackage[utf8]{inputenc} 
\usepackage[T1]{fontenc}    
\usepackage{hyperref}       
\usepackage{url}            
\usepackage{booktabs}       
\usepackage{nicefrac}       
\usepackage{microtype}      

\usepackage{pdfpages} 
\usepackage{colortbl} 
\usepackage{subcaption}
\usepackage{soulutf8}
\usepackage{color}
\usepackage{parskip}

\usepackage{graphicx}
\usepackage{textcomp}
\usepackage{caption}
\usepackage{xcolor}

\usepackage{algorithm}
\usepackage[noend]{algpseudocode}

\usepackage{multirow}
\usepackage{lipsum,booktabs}

\usepackage[normalem]{ulem}

\usepackage{mathtools}

\usepackage{amssymb,amsmath,amsthm}
\usepackage{amsfonts}

\DeclareSymbolFont{cmsymbols}{OMS}{cmsy}{m}{n}
\SetSymbolFont{cmsymbols}{bold}{OMS}{cmsy}{b}{n}
\DeclareSymbolFontAlphabet{\mathcal}{cmsymbols}

\newcommand{\hm}{HiPer Movelets}

\newcommand{\HM}{HiPerMovelets}
\newcommand{\HML}{\HM{}-Log}
\newcommand{\HP}{\HM{}-Pivots}
\newcommand{\HPL}{\HP{}-Log}

\newcommand{\M}{Movelets}

\newcommand{\MM}{MASTERMovelets}
\newcommand{\MML}{\MM{}-Log}

\theoremstyle{definition}
\newtheorem{definition}{Definition}

\DeclareMathOperator{\Tparam}{0.5}
\DeclareMathOperator{\Tparamper}{50\%}

\newcommand{\bl}[1]{\textbf{#1}}
\newcommand{\un}[1]{\underline{#1}}

\newcommand{\PreserveBackslash}[1]{\let\temp=\\#1\let\\=\temp}
\newcolumntype{C}[1]{>{\PreserveBackslash\centering}p{#1}}
\newcolumntype{L}[1]{>{\PreserveBackslash\raggedright}p{#1}}
\newcolumntype{R}[1]{>{\PreserveBackslash\raggedleft}p{.15\textwidth}|}

\title{Fast Discovery of Heterogeneous Subsequences for Fast Trajectory Classification}

\author{
  Tarlis Tortelli Portela, Jonata Tyska Carvalho, Vania Bogorny\\
  Department of Informatics and Statistics\\
  Federal University of Santa Catarina\\
  Florianópolis, Santa Catarina, Brazil\\
  \texttt{tarlis.portela@posgrad.ufsc.br, jonata.tyska@ufsc.br, Vania.Bogorny@ufsc.br}
}

\begin{document}
\maketitle


\keywords{Data Mining \and Multiple Aspect Trajectories \and Hiper Movelets \and Trajectory Classification, Sequence Classification}

\maketitle

\newcommand*{\definitionautorefname}{Definition}
\renewcommand*{\subsectionautorefname}{Section}
\newcommand*{\algorithmautorefname}{Algorithm}

\section{Introduction}\label{sec:intro}

Trajectory classification can be defined as the technique of discovering the class label of a moving object based on its trajectories \cite{Lee2008}. 
It is important for several application domains as, for instance, identifying the transportation mode of a moving object \cite{Etemad2018}, the strength level of a hurricane \cite{Lee2008}, a vessel type (cargo, fish, tourism, etc) \cite{Lee2008}, the owner user of a trajectory \cite{MASTERMov}, etc. 
A recent method for trajectory classification is \MM{}, developed specifically for multiple aspect trajectories \cite{MASTERMov}. 
This method extracts \emph{movelets}, which are subtrajectories that better discriminate each class, and are used as input to  traditional classifiers. 
 \MM{} automatically explores all possible subtrajectories (or subsequences) of any size (e.g. one point, two points, three points, etc.) and explores all dimension combinations (e.g. space; space and time; space, time and POI category, etc), while seeking the best ones to represent the classes. The method  Movelets\cite{Movelets} outperformed all previous works on raw trajectory classification.



The \textit{movelets} are in general subtrajectories that are frequent in trajectories of one class and infrequent in trajectories of other classes, representing a certain behavior, or a routine.
The quality of a movelet is computed based on the distance of a subtrajectory, called movelet candidate, to all subtrajectories of the same size (same number of points) in every class of the database.
In real datasets, the combinatorial explosion of \MM{} makes it very time consuming, but it outperformed all previous methods for trajectory classification in terms of accuracy. 

The \MM{} generates subtrajectories of all sizes (e.g. one point, two points, three points, etc) and dimension combinations, seeking for the best movelet candidates, meaning that for each subtrajectory there is an explosion of dimension combinations.
This means that in a trajectory with ten points and three dimensions, will be generated $55$ subtrajectories, that with $7$ possible dimension combinations will result in $385$ \textit{movelet candidates}. 
The problem is that with increasing dimensionality these numbers exponentially grow as, for instance, a trajectory with $20$ points and $6$ dimensions will result in $210$ subtrajectories, and $13,230$ \textit{movelet candidates}.
It will try $63$ dimension combinations for each subtrajectory, as the number of combinations is given by $2^l-1$, where $l$ is the number of dimensions. 
However, \MML{} can limit subtrajectories to the natural logarithm of its size. 
In the example, the trajectory with $20$ points and $6$ dimensions, \MML{} reduces to $74$ subtrajectories and $4,662$ \textit{movelet candidates}.
Although this strategy significantly reduces the computational cost, it still  computes the distances of each subtrajectory  to all subtrajectories of the entire dataset for comparisons, and generates a large number of movelets to use as input to classifiers, resulting in the curse of dimensionality problem.

Considering the \MM{} high computational cost, in this work we propose a new method for discovering the movelets. 
We claim that frequent subtrajectories in a class will discriminate classes, as human behaviors commonly display recurrent movement patterns, which are more frequent for an object and less frequent for others.

\section{Related Works} \label{sec:related_works}



The algorithms \M{} \cite{Movelets} and \MM{} \cite{MASTERMov} discover \emph{relevant subtrajectories}, called movelets, without the need of extracting other features. 
The \MM{} achieves a much better accuracy than \M{}.
While \M{} encapsulates the distances of all trajectory dimensions in a single distance value, the \MM{} keep distances in a vector using all data dimensions with the best combination.
Ferrero et al. (2020) \cite{MASTERMov} also proposed \MML{}, which reduces the running time of the \MM{} by limiting the size of the movelet candidates to the natural logarithm of the trajectory size.
However, the processing time and computational cost are still extremely high. 
\MM{} \cite{MASTERMov} was developed for multiple aspect trajectory classification, specifically, and largely outperformed state of the art methods in terms of accuracy.
Moreover, the \textit{movelets} are interpretable results, giving insights about the data. 
It is important for researchers looking for vaccines, for instance, that the classification results can be explainable and understandable.


The work of \cite{Vicenzi2020} can perform much faster than \M{} \cite{Movelets}, with similar classification accuracy for a single dimension. 
To use several dimensions, the user must manually test and select the best dimensions combination.
While, \MM{} automatically selects the best dimension combinations with heterogeneous and different numbers of dimensions to generate \textit{movelets}.

MARC, proposed by \cite{MayPetry2020}, is a recent work that uses word embeddings \cite{Mikolov2013}. 
It encapsulates all trajectory dimensions: space, time and semantics, and use them as input to a neural network classifier.
It is the first work to use the geoHash \cite{Geohash2008} on the spatial dimension, combined with other dimensions.
MARC outperforms the \M{} \cite{Movelets} with very high accuracy when using all dimensions, as \M{} is unable to choose the best dimensions. 
However, the resulting patterns of MARC are not interpretable, as the classifier is limited to neural networks.

Some works for trajectory classification were developed for solving a single problem, as transportation mode inference \cite{Etemad2018,Dabiri2018,Xiao2017}, while the works of
\cite{Santos2011,Movelets,MayPetry2020,MASTERMov} were developed for general purposes, such as level of hurricane, animal types, transportation modes, trajectory users, etc. 

Since \M{} and \MM{} are inspired by shapelets, we included works in time series optimization.
Several optimizations are used in these works, as for instance, early abandoning \cite{Ye2011,Mueen2011}, SAX to reduction of search space and run time
\cite{Rakthanmanon2013}, and sampling the dataset for fast candidate search \cite{Ji2019}.
The early abandoning technique is not useful for movelets because multiple aspect trajectories are in general more sparse than time series, and this technique is interesting for sequences with high number of points.

Other works used statistical analysis, such as an Analysis of Variance A(NOVA) metric employed by Zuo et. al. \cite{Zuo2018}.
In time series numerical values, statistical analysis can be used to identify regions from which to extract shapelets that have discriminant power over other classes.
It is also used to filter redundant and low-quality candidates.
Another work, proposed by Zhang et. al. \cite{Zhang2018}, used Local Fisher Discriminant Analysis (LFDA) to determine the most relevant parts of the time series from which to extract shapelet candidates that are most likely to achieve good results.
The works in time series, as the above mentioned, generally search discriminant shapelets in one dimension individually, while \MM{} evaluates the dimensions individually and combined with others.

\section{HiPerMovelets: A method for High Performance Movelet Extraction} \label{cap:method}

In this section we propose a new method for reducing the search space for finding \textit{movelets}, called \HM{}. The method is inspired by the greedy search algorithm for feature selection proposed in \cite{Gigli2016}, as an optimization for \MM{} \cite{MASTERMov}.
Our method is class-aware, and deals with one class at a time instead of all the classes in the dataset at once, thus reducing the search space and time for finding the \textit{movelets}.

\subsection{Main Definitions}\label{sec:main_def}

\textit{Movelet candidates} in \cite{MASTERMov} represent subtrajectories with different sizes and any dimension combination. 
As we propose to reduce the search space for movelets discovery, we define a different type of movelet candidate, described in \autoref{def:hiper-candidate}. 
In our method, a movelet candidate is a subtrajectory that is frequent in the trajectories of a class.

\begin{definition}\label{def:hiper-candidate}
\textit{Movelet  Candidate.} 
A movelet candidate is a tuple $\mathcal{M} = (T_i,s_{(start,end)}, C, \mathbb{W}, L, \textbf{T}^c)$, where $T_i$ is a trajectory from  $\textbf{T'}$; 
$s_{start,end}$ is a subtrajectory extracted from $T_i$;
$C$ is a subset with the candidate dimensions, and $C \subseteq D$;
$\mathbb{W}$ is a set of pairs $(W^s_{T_j}, class_{T_j})$, where $W^s_{T_j}$ is a distance vector that contains the distances of the best alignment of $\mathcal{M}$ to every trajectory $\textbf{T'}$, the same class as ${T_j}$.
The distances are calculated using the best alignment between $s_{(start,end)}$ and each trajectory ${T_j}$ in the dimensions $C$;
$\textbf{T}^c$ are the \textit{covered trajectories}, i.e., a subset of class trajectories that contain the candidate $\mathcal{M}$ ($\textbf{T}^c \subseteq \textbf{T'}$);
$L$ is a pair $(sp, quality)$, where $sp$ is a set of distances $W^s_{T_j}$, with one value for each dimension, that better divide the classes (used to measure the candidate relevance), and \textit{quality} is a relative frequency score given to the candidate $\mathcal{M}$.
\end{definition}
 
The quality score of a movelet candidate is based on the relative frequency (a proportion) that it appears in its class trajectories. 
For each dimension of a movelet candidate, we define a relative frequency as presented in \autoref{eq:aspect_proportion}. 
It describes the relative frequency function for a dimension $k$ ($freq_{k}$) from a \textit{movelet candidate} $\mathcal{M}_j$ to the trajectories $\textbf{T'}$ of a class:

\begin{equation} \label{eq:aspect_proportion}
    freq_{k} =
    \frac{\sum^{i=|\textbf{T'}|}_{i=1} \left ( max({W^*[k]}) - W^s_{T_i}[k] \right )}{max({W^*[k]}) . |\textbf{T'}|}
\end{equation}

Where $W^*$ is a multidimensional vector of trajectory distances from extracted candidates to trajectories of the class ($\textbf{T'}$), and $W^s_{T_i}[k]$ is the distance value from the \textit{movelet candidate} to the trajectory $T_i \in \textbf{T'}$ in the dimension $k \in C$.
To make the proportion more accurate, we normalize the distance value by subtracting it from the maximum distance value ($max({W^*[k]})$) in that dimension.
The frequency is the average between the sum of the distances and the maximum possible sum of distances ($max({W^*[k]}) . |\textbf{T'}|$).
The quality of a movelet candidate (\autoref{eq:quality_proportion}) is the average proportion that a candidate has in trajectories $\textbf{T'}$ of its class in each dimension $k$. 

\begin{equation} \label{eq:quality_proportion}
        \textit{Quality} = \frac{\sum^{k=|C|}_{k=1}freq_k(\mathcal{M}_j, \textbf{T'})}{|C|}
\end{equation}

As there are different dimension combinations for a candidate, we use a ``\textit{soft}'' approach for measuring its frequency and a ``\textit{hard}'' approach to do \textit{trajectory pruning} (\autoref{sec:met-coverage}).
\autoref{eq:aspect_proportion} considers each dimension separately, and than aggregates the results in one quality value (\autoref{eq:quality_proportion}).
he $freq_{k}$ is a weighted average of all distances in each dimension.
This way, even in cases where movelet candidates are not exact matches (when distances are all zeros), the method can still find the most similar ones because the quality is proportional to the maximum distances.

Movelet candidates are selected based on their quality, and only those with highest relative frequency will  become movelets.
Our method  evaluates the movelet candidates with at least $\tau$ relative frequency ($\mathcal{M}.L.quality \geq \tau$), where $\tau$ is calculated from the movelet candidate with highest score.



Our method limits the search for movelets in the set of movelet candidates with high relative frequency. 
A movelet, however, is qualified with \textit{F-Score} considering all trajectories in the dataset, similar to \MM{}.
We define movelets (\autoref{def:movelet2}) as the movelet candidates that have best \textit{F-Score} and no overlapping points.

\begin{definition}\label{def:movelet2}
\textit{Movelet.} 
Given a trajectory $T_i$, and a movelet candidate $\mathcal{M}_x = (T_i,s_{(start,end)}, C, \mathbb{W},$ $L, \textbf{T}^c)$ containing a subtrajectory s with $s_{start,end} \subseteq T_i$, $\mathcal{M}_x$ is a movelet if for each movelet candidate $\mathcal{M}_y$ containing a subtrajectory $u_{f,g}$ with $u_{f,g} \subseteq T_i$ that overlaps $s_{start,end}$ in at least one element, $\mathcal{M}_x.quality > \mathcal{M}_y.quality$.
\end{definition}

In the work of \cite{Gigli2016} for feature selection, features are iteratively chosen as they maximize the relevance for classification, which are the most different ones from the previous chosen feature.
It is a greedy approach to search the most discriminant features without the need to explore the entire dataset.
In a trajectory dataset, trajectories that contain the same movelets are similar and therefore do not need to be analysed.
We introduce the concept of \textit{covered trajectory} in \autoref{def:covered-trajectory} as a criterion for our proposed greedy search algorithm.



\begin{definition}\label{def:covered-trajectory}
\textit{Covered Trajectory.} A trajectory $T_b$ is covered by a movelet candidate of a trajectory $T_a$ when they have the same class and at least half of the movelet candidates of $T_a$ have distance equal to zero to subtrajectories of the same size of $T_b$.~
\end{definition}%

The movelet candidates of a trajectory that cover another trajectory of the same class are likely to represent both trajectories.
Therefore, it is unnecessary to search for movelets in the covered trajectory.
Covered trajectories are then ignored by our method, thus reducing the search space and avoiding to search for movelets in trajectories of the same class.


\subsubsection{\HM{} General Overview}\label{sec:HM}

Our approach, called \HM{} (HIgh PERformance Movelets), extracts movelets that better discriminate classes without searching the entire dataset.

\autoref{alg:hiper_method} summarizes the \HM{} method.
The input  is a set of trajectories $\textbf{T}$ from the dataset and the output is the set of \textit{movelets} $\textbf{M}$ extracted from each class. 
The first step searches for relevant subtrajectories inside the trajectories of each class (line 2). 
Next, it fills a queue with the trajectories $\textbf{T'}$ of one class (line 3) and extracts movelet candidates of one trajectory at a time.
It extracts all possible subtrajectories and selects movelet candidates that have relative proportion higher than $\tau$ (line 6). 

\begin{algorithm}
\caption{\HM{}}\label{alg:hiper_method}
\begin{algorithmic}[1]
\Statex{\textbf{Input}: \textbf{T}// Trajectory dataset}
\Statex{\textbf{Output}: \textbf{M} // set of \textit{movelets}}
\State $\textbf{M} \gets \emptyset$
\For{\textbf{each} \textit{class trajectories} \textbf{T'} \textbf{in} \textbf{T}}
    \State $queue \gets \textbf{T'}$
    \While{\textit{queue} \textbf{not} \{\}}
        \State $T_i \gets queue.pop()$
        \State $\textit{bestCandidates}_{T_i},\; \textit{bucket} \gets hiperExtraction(T_i, \textbf{T'})$ 
        \State $\textbf{M}^{T_i} \gets moveletDiscovery(bestCandidates_{T_i},$ \textbf{T}$)$
        
        \If{$\textbf{M}^{T_i} \; = \; \emptyset$}
            \State \textit{do $moveletDiscovery$ with $10\%$ slices from $bucket$}
        \EndIf
        
        \State $\textit{covered}_{T_i} \gets coveredTrajectories(\textbf{M}^{T_i})$
        \State $queue \gets queue - \textit{covered}_{T_i}$
        \State $\textbf{M} \gets \textbf{M} \cup \textbf{M}^{T_i}$
    \EndWhile
\EndFor
\State $\Return\ \textbf{M}$
\end{algorithmic}
\end{algorithm}

At this step, candidates are compared only to trajectories of the same class, as detailed in \autoref{sec:met-step1}.
Then, in the movelets discovery step, it compares the earlier selected candidates to all trajectories in the dataset with \textit{F-Score} quality (line 7), and the most discriminant are the resulting movelets (\autoref{sec:met-step2}).
However, if \HM{} is not able to discover movelets (line 8), it will start recovering the previous excluded candidates (line 9) with small portions of 10\% of the $bucket$ set at a time, so it will fast search top-bottom from the better qualified candidates to the less frequent and stop when movelets are found. 
Finally, it analyzes the covered trajectories of each movelet of $T_i$, and removes them from the queue (lines 10 and 11), as detailed in \autoref{sec:met-coverage}.

The \HM{} can be divided in three main steps: (i) the hiper movelet candidate discovery, (ii) individual trajectory \textit{movelet} discovery, and (iii) the class trajectory pruning.

\subsubsection{\HM{} Candidate Extraction} \label{sec:met-step1}

The movelet candidates extraction is similar to the \MM{} method, with the difference that \HM{} compares the candidates only with trajectories in the same class, and their quality is based on the frequency inside the class.
As an example of extraction of movelet candidates, \HM{} generates all subtrajectories of $T_i$ and only computes distances to trajectories $\textbf{T'}$ of the class.
Although there is a combinatorial explosion, in this step, it is limited to the number of the class trajectories. 
It qualifies each \textit{movelet candidate}, and filters them by \textbf{low quality} and \textbf{redundancy}.
First, the movelet candidates are filtered by the value of $\tau$, which is learned for each trajectory, and consists of $\Tparamper$ of the relative frequency from the best qualified candidate.
The movelet candidates with $\tau$ relative frequency or higher are compared to the dataset $\textbf{T}$ for discovering the \textit{movelets} (\autoref{sec:met-step2}).
The second filter is to remove non redundant \textit{movelet candidates}.
For instance, if two \textit{movelet candidates} have the same values in at least one dimension, they will remain.
The candidates that repeat in a trajectory are better to describe the movement.
Such filter can greatly reduce the number of candidates keeping the most probable to became movelets.



It is less likely to a \textit{movelet candidate} having a perfect fit (when all distances are zero), as the subtrajectory size increases.
Therefore, we propose two variations for hiper movelet candidates extraction: (i) without pivots, and (ii) using pivots.

The first way of \HM{} candidates extraction, presented in \autoref{alg:selectcan}, generates all \textit{movelet candidates} of any size (or log limit) and than prunes by relative frequency. 
Starting from size one to the limit size $m$ (line 3), it generates all movelet candidates (line 4).
Then, it calculates the relative frequency to each candidate (line 7), so the quality is calculated, and they are sorted.
The $\tau$ value is defined with the first candidate relative frequency (line 9), as they are sorted in descending quality order. 
Lastly, the algorithm filters movelet candidates (line 10) with less than $\tau$ quality.

\begin{algorithm}[h]
\caption{hiperExtraction // Without Pivots}\label{alg:selectcan}
\begin{algorithmic}[1]
\Statex{\textbf{Input}: $\textit{T}_i$ // a trajectory,}
\Statex{\hskip3em $\textbf{T'}$ // Trajectory set for the class,}
\Statex{\textbf{Output}: \textit{bestCandidates} // set of best \textit{hiper candidates}}
\Statex{\hskip3.6em \textit{bucket} // excluded \textit{movelet candidates}}
\State $\textit{candidates}_{T_i} \gets \emptyset$
\State $m \gets \textit{T}_i.length$ \hspace{4em} // OR $\log_2 \left( \textit{T}_i.length \right)$
\For{size \textbf{in} $(1, m)$}
    \State $\textit{candidates}_{size} \gets findCandidates(T_i, \emptyset, size)$
    \State $\textit{candidates}_{T_i} \gets \textit{candidates}_{T_i} \cup \textit{candidates}_{size}$
\EndFor

\For{each $\mathcal{M} \; \textbf{in} \; \textit{candidates}_{T_i}$}
    \State $\mathcal{M}.L.quality \gets quality_{freq}(\mathcal{M}, \textbf{T'})$
\EndFor
\State $\textit{candidates}_{T_i}.sort()$

\State $\tau \gets \textit{candidates}_{T_i}[0].L.quality \; * \; \Tparam$
\State $\textit{bestCandidates} \gets moveletCandidatePruning(\textit{candidates}_{T_i}, \textbf{T'}, \tau)$

\State $\textit{bucket} \gets \textit{candidates}_{T_i} - \textit{bestCandidates}$

\State $\Return\ \textit{bestCandidates}, \; \textit{bucket}$
\end{algorithmic}
\end{algorithm}

\begin{figure}[]
\centering
\includegraphics[width=0.5\textwidth]{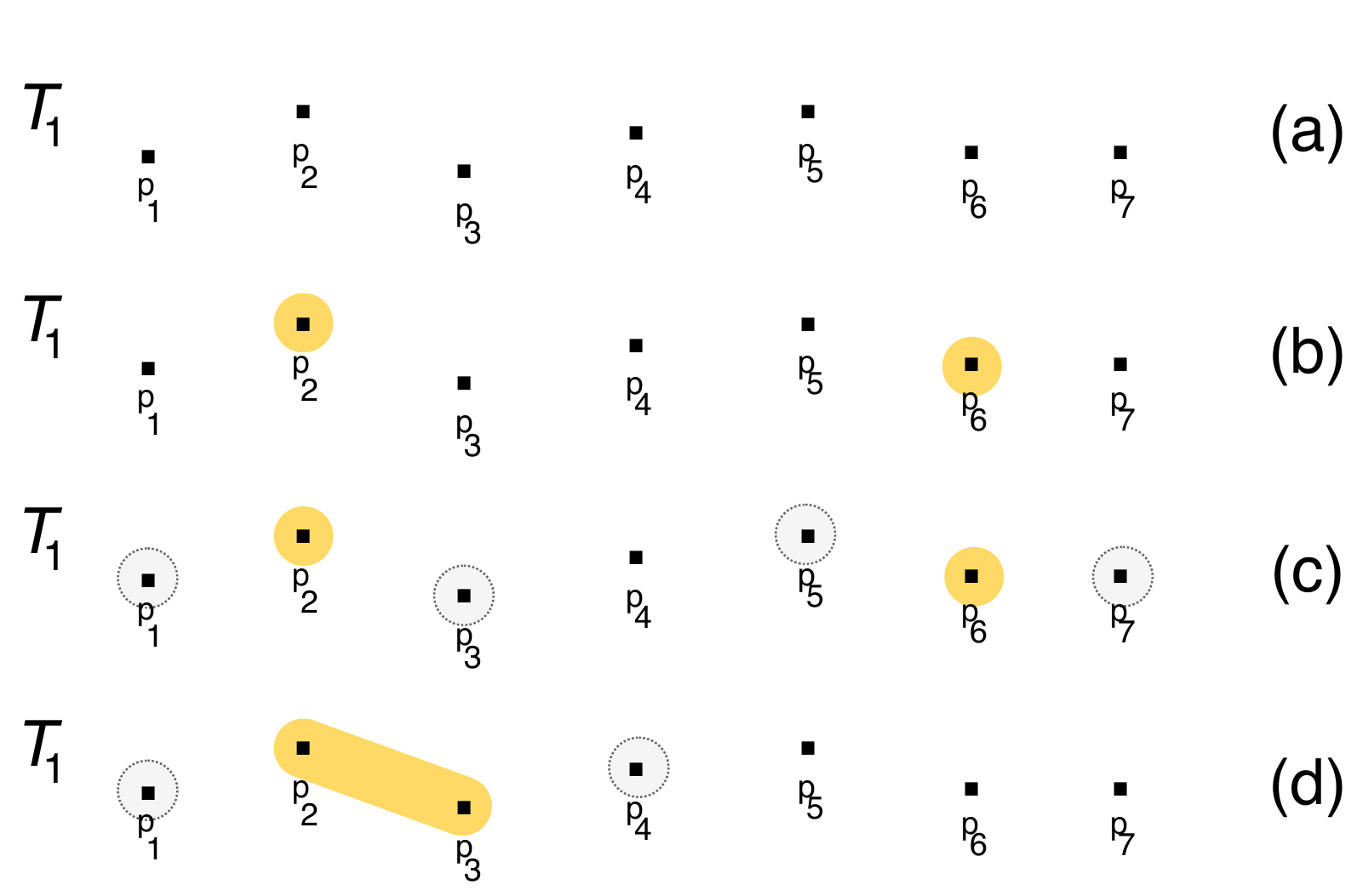}
\caption{(a) Trajectory $T_1$, (b) The pivots of size one, (c) The pivots neighborhood, and (d) one pivot of size two neighborhood.}
\label{fig:pivots}
\end{figure}

The second strategy is exemplified in \autoref{fig:pivots}, that shows a trajectory with seven points (\autoref{fig:pivots} a). It finds \textit{movelet candidates} of size one to the limit size, in each step pruning the candidates by relative frequency.
The first time, it generates the candidates of size one and filters them by relative frequency \autoref{fig:pivots} (b) (suppose this points are $p_2$ and $p_6$.)
Subsequently, it looks for candidates of size two in the nearest neighbors \autoref{fig:pivots} (c), either starting or ending at the points of the previous candidates of size one (the neighbor points $p_1$, $p_3$, $p_5$, and $p_7$).
It generates the size two candidates ($p_1 \rightarrow p_2$, $p_2 \rightarrow p_3$, $p_5 \rightarrow p_6$, and $p_6 \rightarrow p_7$), prunes again by low relative frequency (candidate $p_2 \rightarrow p_3$ in \autoref{fig:pivots} d), and searches new candidates of size three in the neighborhood points ($p_1$ and $p_4$).
Each candidate will generate two new candidates, one added with the previous point and the other added with the next point, similar to the idea of nearest neighbour.
However, our method removes candidates with overlapping points, thus keeping the candidate with higher relative frequency.

\autoref{alg:selectcan_wp} presents the pivots strategy.
First, it generates the candidates of size one (line 6), assess their relative frequency (line 8), sort in descending quality order (line 9) and than filter the candidates relative frequency inferior of $\tau$ (line 11).
Then, it looks for candidates of size two from the neighborhood of the previous candidates of size one (line 6), and prunes again by low relative frequency (line 11).
The \texttt{findCandidates} function (line 6) will generate subtrajectories from the previous size candidates adding the neighbor point in the beginning or end.
It goes on increasing the size until it reaches the trajectory size limit (line 5).
\begin{algorithm}[h]
\caption{hiperExtraction // With Pivots}\label{alg:selectcan_wp}
\begin{algorithmic}[1]
\Statex{\textbf{Input}: $\textit{T}_k$ // a trajectory,}
\Statex{\hskip3em $\textbf{T'}$ // Trajectory set for the class,}
\Statex{\textbf{Output}: \textit{bestCandidates} // set of best \textit{hiper candidates}}
\Statex{\hskip3.5em $\textit{bucket}$ // set of the remaining movelet candidates}
\State $\textit{bestCandidates} \gets \emptyset$
\State $\textit{candidates}_{size} \gets \emptyset$
\State $\textit{bucket} \gets \emptyset$
\State $m \gets \textit{T}_i.length$ \hspace{4em} // OR $\log_2 \left( \textit{T}_i.length \right)$
\For{size \textbf{in} $(1, m)$}
    \State $\textit{candidates}_{T_i} \gets findCandidates(T_i, \textit{candidates}_{size}, size)$
    
    \For{each $\mathcal{M} \; \textbf{in} \; \textit{candidates}_{T_i}$}
        \State $\mathcal{M}.L.quality \gets quality_{freq}(\mathcal{M}, \textbf{T'})$
    \EndFor
    \State $\textit{candidates}_{T_i}.sort()$
    
    \State $\tau \gets \textit{candidates}_{T_i}[0].L.quality \; * \; \Tparam$
    \State $\textit{candidates}_{size} \gets moveletCandidatePruning(\textit{candidates}_{T_i}, \textbf{T'}, \tau)$
    
    \State $\textit{bestCandidates} \gets \textit{bestCandidates} \cup \textit{candidates}_{size}$
    \State $\textit{bucket} \gets \textit{bucket} \cup (\textit{candidates}_{T_i} - \textit{candidates}_{size})$
\EndFor
\State $\Return\ \textit{bestCandidates}, \; \textit{bucket}$
\end{algorithmic}
\end{algorithm}

When a dataset has more dimensions, more patterns can be found and it is easier to find discriminant movelets.
Indeed, the more is information available with combined dimensions, the more unique behaviours can be found.
For instance, datasets with one dimension are harder for finding discriminating patterns in classes and \HM{} might end not finding movelets among the better qualified candidates.
\HM{} keeps the set of removed candidates($bucket$ in \autoref{alg:selectcan_wp} line 13), and it will use the $bucket$ set to look for movelets in case none was previously found.

\subsubsection{Movelets Discovery}\label{sec:met-step2}

The movelets discovery is similar to \MM{} algorithm \cite{MASTERMov}, it computes and filters the movelet candidates using the \textit{F-Score} quality.
The difference is that our method only calculates this quality for the class-frequent movelet candidates.
To measure the \textit{F-Score} is necessary to compare each candidate with the entire dataset, and by reducing the number of movelet candidates, this task can be done significantly faster.

This step is represented by the subroutine \texttt{moveletDiscovery} in \autoref{alg:hiper_method} (line 7), in which the movelets are the best representations of the behaviors of a class.
First, it filters candidates that are equal, what can happen when a behavior repeats in the same trajectory.
Then, \HM{} calculates distances for all movelet candidates to the dataset trajectories, and assesses their \textit{F-Score} quality.
Then, candidates with zero quality and overlapping points are discarded.
If the resulting movelets are an empty set, the \HM{} will repeat this step with the $bucket$ set of candidates previously excluded.
Only the \textit{movelet candidates} with best \emph{quality} and without point overlapping in the trajectory are kept and called \textit{movelets}.

\subsubsection{Trajectory Pruning}\label{sec:met-coverage}

From the discovered \textit{movelets} of each trajectory, \HM{} evaluates the covered trajectories. 
\autoref{alg:covered} shows how our method finds covered trajectories from the movelets of trajectory $T_i$ (line 3).
First, each trajectory is mapped with the number of occurrences in the movelets coverage (lines 4 to 5).
Then, it selects the trajectories covered by at least half of the movelets (lines 6 to 8).
If a class contains similar trajectories, than \HM{} search space can be reduced as the movelets will cover more examples (trajectories) in the class. 
However, classes with distant trajectories will need more examples to discover \textit{movelets}.

\begin{algorithm}[]
\caption{coveredTrajectories}\label{alg:covered}
\begin{algorithmic}[1]
\Statex{\textbf{Input}: $\textbf{M}^{T_i}$ // set of \textit{movelets} of trajectory $T_i$}
\Statex{\textbf{Output}: \textit{covered} // set of \textit{covered trajectories}}
\State $\textit{covered} \gets \emptyset$
\State $covered\_count \gets Map[]$
\For{each $\textit{movelet}$ \textbf{in} $\textbf{M}^{T_i}$}
    \For{each $\textit{T}_x$ \textbf{in} $\textit{movelet}.coveredTrajectories$}
        \State $covered\_count[\textit{T}_x]++$
    \EndFor
\EndFor
\For{each $\textit{T}_y$ \textbf{in} $covered\_count$}
    \If {$covered\_count[\textit{T}_y]> (|\textbf{M}^{T_i}| / 2)$}
    \State $\textit{covered} \gets \textit{covered} \; \cup \;  \textit{T}_y$
    \EndIf
\EndFor

\State $\Return\ \textit{covered}$
\end{algorithmic}
\end{algorithm}

\section{Experimental Evaluation}\label{cap:experiments}

We evaluated the \hm{} approach by comparing their computational cost and classification power. 
First, we evaluated their computational cost in learning the model, and the classification power with \MML{}.

\subsection{Datasets} \label{sub:datasets}

We performed preliminary experiments with the datasets of multiple aspect trajectories to evaluate used in \cite{MASTERMov,MayPetry2020, CamilaThesis}.
They are composed of check-in trajectories and are enriched with semantic dimensions from three different Location-Based Social Networks (LBSN).
The trajectories were splitted by week in order to increase the trajectories examples of each user as necessary for classification tasks.
The multiple aspect trajectory datasets were used for evaluating the \HM{}, because of their number of dimensions and because \MML{} was tested with these datasets.
We evaluated our method over the same datasets used in \cite{MASTERMov,MayPetry2020}, with the average trajectory size, the number of trajectories, points and classes, and the attributes of the dataset.
We also evaluated this datasets using only the spatial dimension, referred as ``spatial only''.

\index{Paragraph}\textbf{Brightkite:} dataset from Brightkite social media, collected between April 2008 and October 2010 \cite{Cho2011}. 
A total  of $300$ users were randomly selected, with a total of $7,911$ trajectories, a minimum of $10$ points and maximun of $50$ points per trajectory for consistency.
Trajectories provide the anonymized user of the check-in, the POI, space and time information, and were enriched with the semantic weekday information.

\index{Paragraph}\textbf{Gowalla:} dataset from the Gowalla  Location-Based Social Network (LSBN), collected between February 2009 and October 2010 \cite{Cho2011}.
We used a total of $300$ random users for analysis, limiting the trajectory sizes of minimum $10$ and maximun $50$ points, resulting in a total of $5,329$ trajectories.
It is composed by the same dimensions as the Brightkite dataset, including the enriched semantic information of the weekday.

The Foursquare dataset is  from the Foursquare social media, having the user  check-in, the POI and the spatial position in time when the check-in was made.
Trajectories were enriched with semantic information of: (i) the weekday, (ii) the POI from the Foursquare API\footnote{https://developer.foursquare.com/} with the POI category, and the numerical information of its price and rating. 
The trajectories were splitted in weeks to increase the examples for each user with at least $10$ check-ins.

\index{Paragraph}\textbf{Foursquare NYC:} the dataset was collected between April 2012 and February 2013 \cite{Yang2015}. 
It has a total of $227,428$ check-in points of $1,083$ users in New York, USA. 
It provides $193$ users for this dataset. 
The dataset has a total of $3079$ trajectories, with trajectory sizes varying from $10$ to $144$.
The dataset trajectories were enriched with the weather information at the POI location and time with the Weather Wunderground API\footnote{https://www.wunderground.com/weather/api/}.

\subsection{Experimental Setup} \label{sub:setup}

The datasets are split in a hold-out proportion of $70\%$ training and $30\%$ testing, with respect to the class balance.
We evaluated three classification models: the Multilayer-Perceptron (MLP), the Random Forest (RF) and the Support Vector Machine (SVM), as they are commonly used and achieved good results in \cite{MASTERMov,MayPetry2020,CamilaThesis}.
The classifiers were implemented using the Python language, with the \texttt{keras}\footnote{https://keras.io/} package in the same way as \cite{MASTERMov,CamilaThesis}. 
The MLP has a fully-connected hidden layer with $100$ units, a Dropout Layer rate of $0.5$, learning rate of $10^-3$ and Output Layer with softmax activation.
It was also applied Adam Optmization to improve the learning time and to avoid categorical cross entropy loss, same as in the work of \cite{CamilaThesis}, with $200$ of batch size, and a total of $200$ epochs for each training.
The Random Forest was built with $300$ decisions trees.
The Support Vector Machine (SVM) algorithm was implemented with linear kernel, and probability estimates are enabled.

In the \MM{} configurations, the distance was calculated in each dimension by using: (i) euclidean distance for the space, (ii) difference for the numerical, and (iii) simple equality (if is equal or not) for the semantical.
 
The experiments were performed in an
\textit{Intel(R) Core(TM) i7-9700 CPU @ 3.00GHz}, with $8$ cores (limited to $3$ threads) and \textbf{32GB} of intern memory.

\subsection{Preliminary Results} \label{sub:results}

We compared the classification results of the \MML{} and our approaches, by using the Accuracy metric. 
The results for each dataset are presented in \autoref{tab:results_all}, where the higher values for each metric are presented in bold, and the second best are underlined.
MARC was the fastest method in almost all the experiments, but its running time is not compared as \HM{} and \MM{} methods, as they were limited to three threads for a fair time comparison between them.
In addition, MARC had the eight cores and GPU acceleration to run.

In this experiment we compare the number of \textit{movelet candidates} generated by the \MML{} and \HML{}, the number of movelets, the time spent in this task, the accuracy and time of classification in each classifier, and the \HML{} number of trajectories it looked for movelets extraction.
Table \ref{tab:results_all} summarizes the results, where the main observation is that the method \HPL{} is at least $95\%$ faster than the \MML{} in time comparison in any of the specific datasets, and \HML{} is at least $71\%$ faster in the same datasets.
\HPL{} and \HML{} in the spatial only datasets are, respectively, at least $78\%$ and $70\%$ faster than \MML{}.
Note that in the spatial only dataset of Foursquare NYC \MML{} did not output movelets. 

\begin{table}[!htb]
\caption{Results for Brightkite, Gowalla and Foursquare NYC datasets.}
\label{tab:results_all}
\begin{subtable}{.5\linewidth}
\resizebox{\textwidth}{!}{
\centering
\footnotesize
\begin{tabular}{|c|r||R|R|R|R|}
\hline
\hline
Dataset &   & MARC & MASTER. -Log & HiPer. -Log & HiPer. Pivots -Log \\

\hline
\multirow{11}{2cm}{Brightkite (spatial only)}
&     Candidates &         - &   409,380 &        288,319 &        192,811 \\
&       Movelets &         - &    83,422 &    \bl{38,186} &    \un{39,802} \\
\cline{2-6}
&            MLP &\un{94.158}& \bl{94.849} &    94.033 &    93.970 \\
&             RF &                   &      94.095 &    93.342 &    93.530 \\
&            SVM &                   &      93.970 &    93.342 &    93.279 \\
\cline{2-6}
&Extraction Time &      8m48s& 1h16m15s&    \un{22m49s} &    \bl{18m12s} \\
&       Time MLP &           &    36m44s &    \bl{14m38s} &    \un{15m09s} \\
&        Time RF &           &       28s &       \bl{16s} &       \un{17s} \\
&       Time SVM &           &  1h51m36s &    \bl{41m05s} &    \un{43m07s} \\
\cline{2-6}
&Trajs. Compared &         - &         - &     4,549 &     4,617 \\
& Trajs. Pruned  &         - &         - &     1,770 &     1,702 \\

\hline
\hline
\multirow{11}{2cm}{Gowalla (spatial only)}
&     Candidates &         - &   318,326 &    252,281 &    146,825 \\
&       Movelets &         - &    52,711 &\bl{11,860} &\un{12,380} \\
\cline{2-6}
&            MLP &87.723& \bl{94.752} &    90.628 &\un{91.753}\\
&             RF &              &      91.565 &    89.784 &    90.159 \\
&            SVM &              &      91.471 &    88.004 &    87.629 \\
\cline{2-6}
&Extraction Time &      5m58s&  42m54s &     \bl{4m44s} & \un{6m54s} \\
&       Time MLP &           &    12m23s &     \bl{2m52s} & \un{2m59s} \\
&        Time RF &           &  \un{14s} &        \bl{6s} &    \bl{6s} \\
&       Time SVM &           &  1h18m20s &    \bl{12m16s} &\un{12m52s} \\
\cline{2-6}
&Trajs. Compared &         - &         - &     3,381 &     3,345 \\
& Trajs. Pruned  &         - &         - &       881 &       917 \\

\hline
\hline
\multirow{11}{2cm}{Foursquare NYC (spatial only)}
&     Candidates &         - &   240,297 &   181,549 &   108,418 \\
&       Movelets &         - &\textit{0} &\un{7,521} &\bl{7,411} \\
\cline{2-6}
&            MLP &\bl{92.241}&         - &\un{91.552} &    91.034 \\
&             RF &                   &         - &     89.828 &    89.655 \\
&            SVM &                   &         - &     88.966 &    88.621 \\
\cline{2-6}
&Extraction Time &      8m08s&   9m36s & \un{1m53s} &     \bl{1m44s} \\
&       Time MLP &           &         - & \un{1m05s} &     \bl{1m03s} \\
&        Time RF &           &         - &    \un{3s} &        \bl{2s} \\
&       Time SVM &           &         - & \un{2m47s} &     \bl{2m40s} \\
\cline{2-6}
&Trajs. Compared &         - &         - &     1,897 &     1,893 \\
& Trajs. Pruned  &         - &         - &       602 &       606 \\

\hline
\hline
\end{tabular}}
\end{subtable}%
\begin{subtable}{.5\linewidth}
\resizebox{\textwidth}{!}{
\centering
\footnotesize
\begin{tabular}{|c|r||R|R|R|R|}
\hline
\hline
Dataset &   & MARC & MASTER. -Log & HiPer. -Log & HiPer. Pivots -Log \\
\hline

\multirow{11}{2cm}{Brightkite (specific)}
&     Candidates &         - & 6,140,700 &   3,355,050 &    773,364  \\
&       Movelets &         - &    88,763 & \un{34,586} & \bl{33,280} \\
\cline{2-6}
&            MLP &95.477& \bl{96.420} &\un{96.357} &    95.729 \\
&             RF &              &      95.352 &     95.791 &    95.415 \\
&            SVM &              &      95.854 &     95.791 &    95.603 \\
\cline{2-6}
&Extraction Time &       6m26s &  22h48m01s &\un{2h40m07s}&  \bl{36m00s}\\
&       Time MLP &             &     38m40s &  \un{13m15s}&  \bl{12m46s}\\
&        Time RF &             &   \un{22s} &     \bl{12s}&     \bl{12s}\\
&       Time SVM &             &   4h21m54s &\un{1h55m01s}&\bl{1h46m07s}\\
\cline{2-6}
&Trajs. Compared &         - &         - &     3,657 &     3,299 \\
& Trajs. Pruned  &         - &         - &     2,662 &     3,020 \\

\hline
\hline
\multirow{11}{2cm}{Gowalla (specific)}
&     Candidates &         - & 4,774,890 &  3,064,335 &        607,977 \\
&       Movelets &         - &    62,141 & \un{20,312} &    \bl{17,459} \\
\cline{2-6}
&            MLP &\un{95.314}& \bl{95.689} &    94.564 &    94.845 \\
&             RF &                   &      94.283 &    93.440 &    92.784 \\
&            SVM &                   &      92.596 &    88.379 &    89.503 \\
\cline{2-6}
&Extraction Time &     11m13s&   9h23m59s &\un{54m26s} &\bl{14m01s} \\
&         T. MLP &           &     14m35s & \un{4m59s} & \bl{4m16s} \\
&          T. RF &           &        14s &    \un{8s} &    \bl{7s} \\
&         T. SVM &           &   2h00m42s &\un{47m04s} &\bl{38m48s} \\
\cline{2-6}
&Trajs. Compared &         - &         - &     2,765 &     2,318 \\
& Trajs. Pruned  &         - &         - &     1,497 &     1,944 \\

\hline
\hline
\multirow{11}{2cm}{Foursquare NYC (specific)}
&     Candidates &         - & 61,275,735 &  25,697,880 &  3,248,927 \\
&       Movelets &         - &     36,353 & \un{14,956} &\bl{11,552} \\
\cline{2-6}
&            MLP &\bl{98.966}&    97.069 &\un{98.276} &    95.345 \\
&             RF &                   &    95.345 &     96.552 &    96.552 \\
&            SVM &                   &    90.862 &     83.793 &    74.138 \\
\cline{2-6}
&Extraction Time &       9m09s & 29h49m49s &\un{8h32m47s} & \bl{1h28m02s} \\
&         T. MLP &             &     4m50s &.  \un{2m05s} &    \bl{1m37s} \\
&          T. RF &             &   \un{4s} &      \bl{3s} &       \un{4s} \\
&         T. SVM &             &    34m24s &. \un{14m44s} &   \bl{11m15s}\\
\cline{2-6}
&  Trajs. Looked &         - &         - &     1,173 &       674 \\
& Trajs. Pruned  &         - &         - &     1,326 &     1,825 \\

\hline
\hline
\end{tabular}}
\end{subtable}
\end{table}




\MML{} has the highest accuracy in almost all experiments.
In the specific and spatial only datasets, however, our method had almost the same or higher accuracy than \MML{}.

\section{Conclusion} \label{cap:conclusion}
In this draft we proposed a method for fast movelet extraction, that is more efficient than MasterMovelets and keeps a similar accuracy. 
In future works we will perform several experiments with 5-fold cross validation and scalability analysis.
\bibliographystyle{unsrt}  
\bibliography{bibliography/references}

\end{document}